%% file: aaai23.tex
\documentclass[letterpaper]{article} 
\usepackage{aaai23}  
\usepackage{times}  
\usepackage{helvet}  
\usepackage{courier}  
\usepackage[hyphens]{url}  
\usepackage{graphicx} 
\usepackage{natbib}  
\usepackage{caption} 
\usepackage{algorithm}
\usepackage{algorithmic}

\usepackage{newfloat}
\usepackage{listings}
\usepackage{caption}
\usepackage{graphicx}
\usepackage{float} 
\usepackage{subfigure}
\usepackage{amsmath}
\usepackage{mathrsfs}
\usepackage{amsmath}
\usepackage{amsthm}
\usepackage{booktabs}
\usepackage{algorithm}
\usepackage{algorithmic}
\usepackage{multirow}
\usepackage{amssymb}
\usepackage{bbding}
\usepackage{mathrsfs}
\usepackage[normalem]{ulem}
\usepackage{color}
\usepackage[dvipsnames]{xcolor}
\DeclareCaptionStyle{ruled}{labelfont=normalfont,labelsep=colon,strut=off} 
\floatstyle{ruled}
\newfloat{listing}{tb}{lst}{}
\floatname{listing}{Listing}
\title{Reviewing Labels: Label Graph Network with \textit{Top-k} Prediction Set \\ for Relation Extraction}
\author{
Bo Li$^{1,2}$, 
Wei Ye$^{1}$\thanks{{} {} Corresponding author.}, 
Jinglei Zhang$^{1,2}$, 
Shikun Zhang$^{1}$$^*$ \\}

\affiliations{
$^1$National Engineering Research Center 
  for Software Engineering, Peking University \\
  $^2$School of Software and Microelectronics, Peking University\\
  {deepblue.lb@stu.pku.edu.cn}, {wye@pku.edu.cn},\\ {jinglei.zhang@stu.pku.edu.cn}, {zhangsk@pku.edu.cn}
}

\usepackage{bibentry}

\begin{document}

\maketitle

\begin{abstract}

The typical way for relation extraction is fine-tuning large pre-trained language models on task-specific datasets, then selecting the label with the highest probability of the output distribution as the final prediction. However, the usage of the \textit{Top-k} prediction set for a given sample is commonly overlooked. In this paper, we first reveal that the \textit{Top-k} prediction set of a given sample contains useful information for predicting the correct label. To effectively utilizes the \textit{Top-k} prediction set, we propose \textbf{L}abel \textbf{G}raph Network with \textit{\textbf{Top-k}} Prediction Set, termed as \textbf{KLG}. Specifically, for a given sample, we build a label graph to review candidate labels in the \textit{Top-k} prediction set and learn the connections between them. We also design a dynamic $k$-selection mechanism to learn more powerful and discriminative relation representation. Our experiments show that \textbf{KLG} achieves the best performances on three relation extraction datasets. Moreover, we observe that \textbf{KLG} is more effective in dealing with long-tailed classes. 

\end{abstract}

\input{1-intro}
\input{3-Method}
\input{4-experiments}
\input{5-analysis}
\input{2-related_work}

\section{Conclusion}
In this paper, we first reveal that the \textit{Top-k} prediction set of a given sample contains helpful information when dealing with relation extraction. Then we propose \textbf{L}abel \textbf{G}raph Network with \textit{\textbf{Top-k}} Prediction Set (\textbf{KLG}), a new model that fully utilizes the \textit{Top-k} prediction set by building a label graph neural network with the dynamic $k$-selection mechanism. By digging the potentially useful information from the \textit{Top-k} prediction set and reviewing these labels, \textbf{KLG} achieves new state-of-the-art results on three relation extraction datasets. Besides, \textbf{KLG} is particularly good at handling long-tailed classes. We hope this research could inspire researchers to further effectively explore the usage of the \textit{Top-k} prediction set.

\bibliography{aaai23}
\%nobibliography{aaai23}


\end{document}

%% file: 1-intro.tex
\section{Introduction}\label{intro}

Relation extraction (RE) is a fundamental natural language processing task with various downstream applications. RE aims to classify the relation between two target entities under a given input. By incorporating powerful pre-trained language models (PLMs) \cite{DBLP:conf/naacl/DevlinCLT19,DBLP:journals/corr/abs-1907-11692}, researchers achieve impressive performances on most of relation extraction datasets \cite{DBLP:conf/acl/SoaresFLK19,DBLP:conf/acl/YeLXSCZ19,DBLP:conf/emnlp/YamadaASTM20,DBLP:journals/corr/abs-2102-01373,DBLP:conf/acl/LyuC21,DBLP:conf/emnlp/RoyP21}. 

Generally, these approaches first fine-tune a PLM on downstream datasets, then get the prediction based on the output probability distribution of the classification layer. Selecting the label with the highest probability as the final prediction is a default solution, however, the \textit{Top-k} prediction set---\textit{Top-k} predictions with the highest probability---is largely overlooked. We argue that the \textit{Top-k} prediction set contains some information that may be useful for the prediction. Specifically, given a well-trained PLM and a sample, if the prediction is wrong, the \textit{Top-k} prediction set may contain the ground truth label. We can review labels in it and correct the prediction. Even if the prediction is correct, we can still establish the label connections between the ground truth and the other labels in the \textit{Top-k} prediction set, and further refine the relation representation. 

\begin{figure}
	\centering
	\subfigure[]{
		\begin{minipage}[t]{0.5\linewidth}
			\centering
			\includegraphics[width=0.99\linewidth]{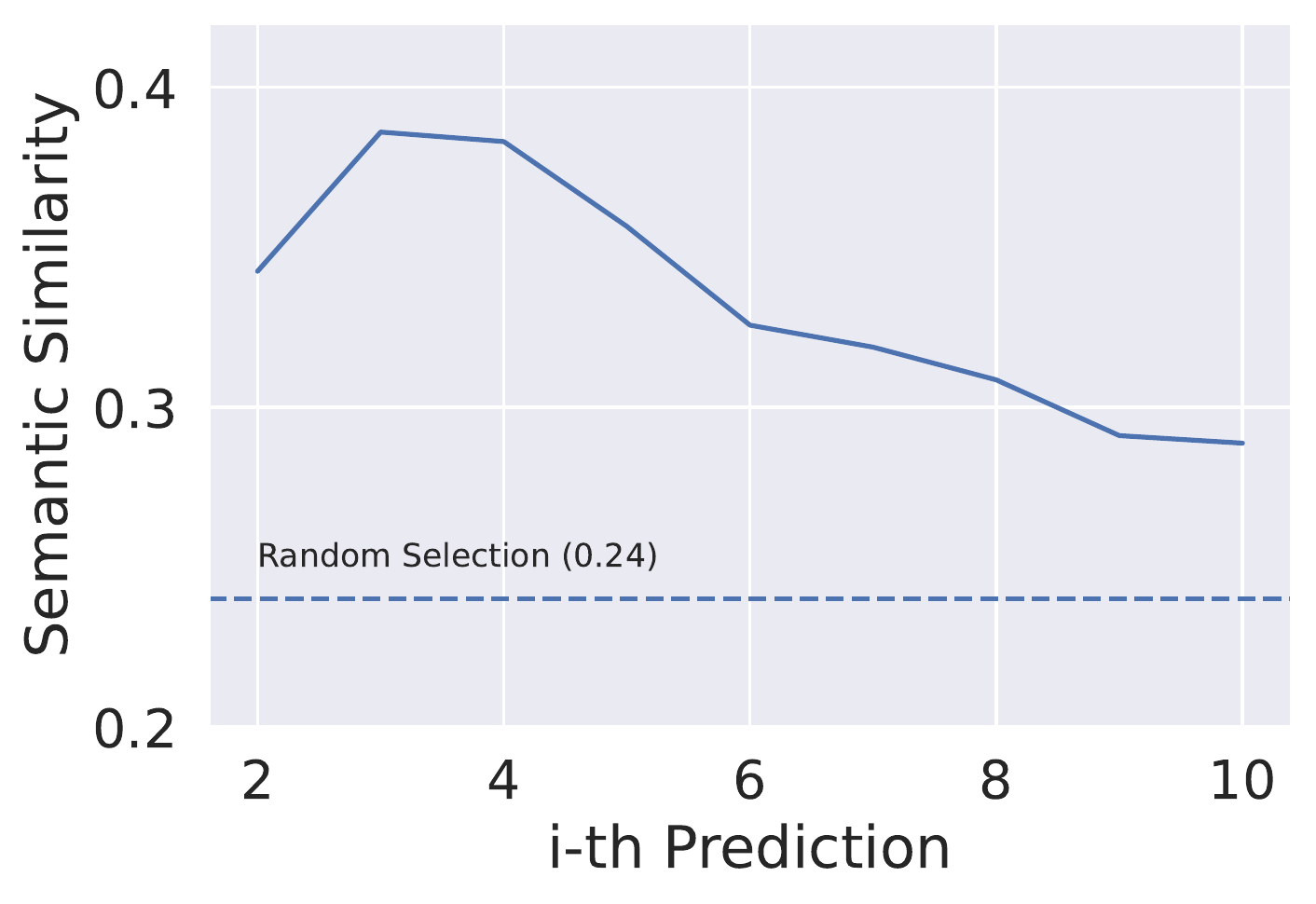}
		\end{minipage}
	}%
	\subfigure[]{
		\begin{minipage}[t]{0.5\linewidth}
			\centering
			\includegraphics[width=0.99\linewidth]{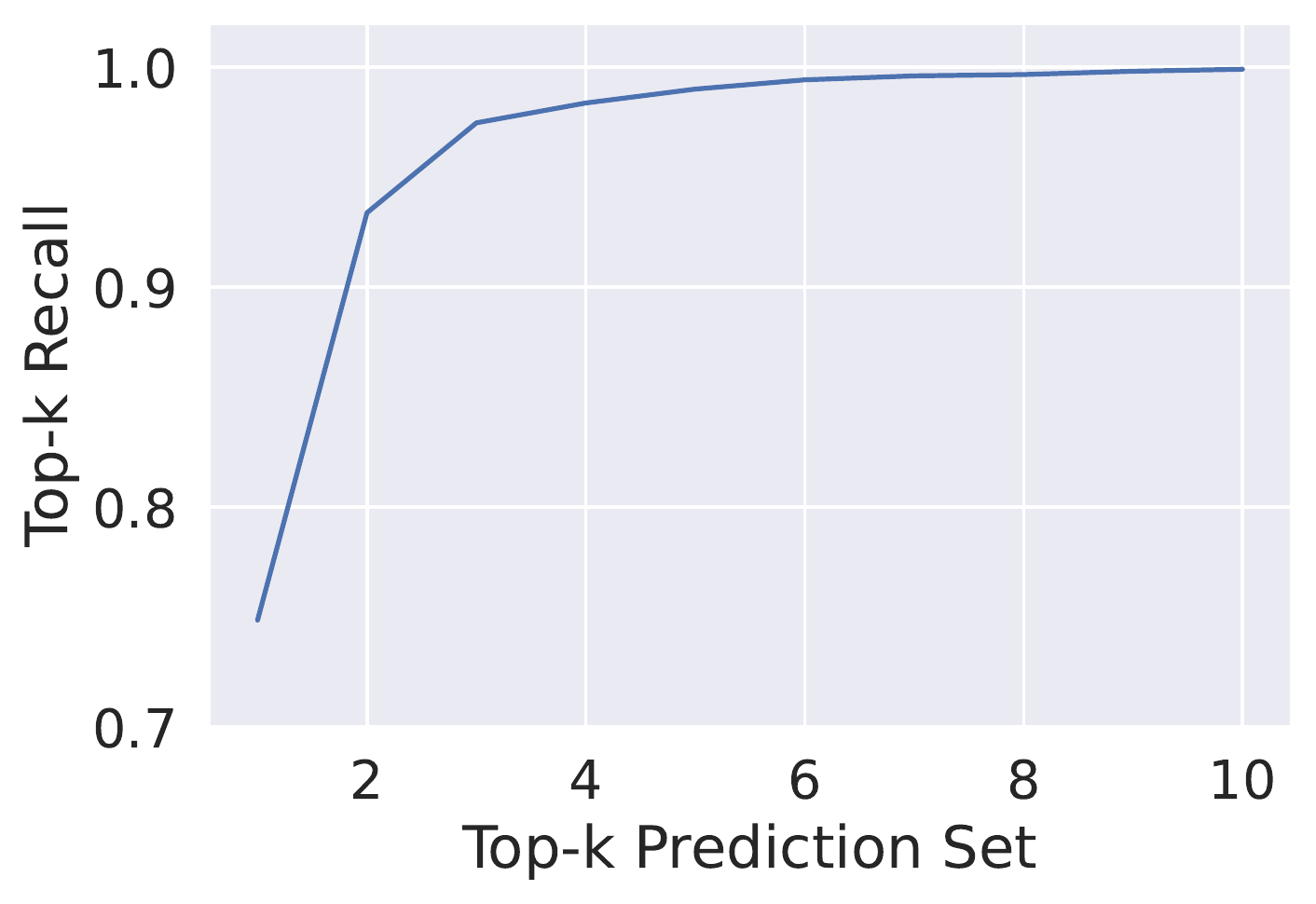}
		\end{minipage}
	}%
	\centering
	\caption{The statistical information of the \textit{Top-k} prediction sets from a well-trained relation extraction model. (a) shows the average semantic similarity between the ground truth label and its $i$-th prediction. (b) shows the recall of different \textit{Top-k} prediction set. We can observe that labels existing in the \textit{Top-k} prediction set has strong connections with the corresponding ground truth label. Besides, the \textit{Top-6} prediction set achieves a recall score of more than 0.99, meaning that most ground truth labels exist in \textit{Top-6} prediction sets.}
	\label{pilot}
\end{figure}

We first conduct a pilot experiment to explore whether \textit{Top-k} prediction sets contain meaningful information. We first train a relation extraction model using RoBERTa-large \cite{DBLP:journals/corr/abs-1907-11692} on the TACRED \cite{DBLP:conf/emnlp/ZhangZCAM17} dataset, and then use this well-trained model to obtain the \textit{Top-k} prediction set of each test sample. The statistical information is visualized in Figure \ref{pilot}. We report the average semantic similarity between the ground truth label and the $i$-th label in the corresponding \textit{Top-k} prediction set.\footnote{We use SBERT\cite{DBLP:conf/emnlp/ReimersG19} to compute the cosine similarity between relation names, since relation names in TACRED are meaningful phases.} For each example, we also compute the semantic similarity between its ground truth and a randomly selected relation as the lower bound, denoted as random selection. We find that labels in the \textit{Top-k} prediction set have strong connections with its ground truth label, while random selection shows much lower similarity. In Figure \ref{pilot} (b), we show the recall of the \textit{Top-k} prediction set with different $k$. We can easily observe that the recall grows fast as the $k$ increases, e.g., \textit{Top-1} recall is only around 0.75, and \textit{Top-6} recall is already larger than 0.99. The above results verify our assumption that the \textit{Top-k} prediction set contains available information, and we may benefit relation extraction if it can be used properly.

Based on the above observations, we propose \textbf{L}abel \textbf{G}raph Network with \textit{\textbf{Top-K}} Prediction Set (\textbf{KLG}), which could utilize the \textit{Top-k} prediction set to improve the performance on relation extraction task. We first fine-tune a PLM on downstream datasets with a standard supervised learning paradigm. This PLM then automatically generates the \textit{Top-k} prediction set for each sample. These \textit{Top-k} prediction sets will be treated as supplementary information to train \textbf{KLG}. Specifically, \textbf{KLG} first builds a label graph, which sets all pre-defined labels as nodes, and extracts the adjacent information from the \textit{Top-k} prediction set. In this way, our model could review candidate labels existing in the \textit{Top-k} prediction set and learn the connections between them. 

However, the performance maybe sensitive to a fixed $k$. Performing a grid search is time-expensive and hinders model generalization. Therefore, we considered bypassing the determination of $k$ and letting the model learn by itself from multiple possible \textit{Top-k} prediction sets (or multiple label graphs). To make the model more robust and learn diverse information from the \textit{Top-k} prediction set, we design a dynamic $k$-selection mechanism. For a given sample and its \textit{Top-k} prediction set, we randomly select different \textit{Top-$\hat{k}$} predictions to form various label graphs. Then a graph contrastive learning loss is used to guide the representations of the same sample from different graphs to be close (positive examples), while the representations of different samples to be apart from each other (negative examples). Our dynamic $k$-selection mechanism makes \textbf{KLG} more robust and further achieves better performances.

To verify the effectiveness of \textbf{KLG}, we conduct extensive experiments on three different relation extraction datasets. We find that \textbf{KLG} brings significant improvements over strong baseline models and achieves better results than previous state-of-the-art methods (\S\ref{main_result}). Besides, we also explore why does \textbf{KLG} work, and the experiments show that \textbf{KLG} has a strong ability to improve the performance on long-tailed classes (\S \ref{long_tail}). 

To sum up, our contributions are as follows:

\begin{enumerate}
    \item We show that labels existing in the \textit{Top-k} prediction set has strong connections with the corresponding ground truth label, which is largely ignored by previous works.
    
    \item We propose \textbf{L}abel \textbf{G}raph Network with \textbf{Top-K} Prediction Set (\textbf{KLG}), which contains a label graph and the dynamic $k$-selection mechanism. \textbf{KLG} could effectively leverage useful information in the \textit{Top-k} prediction set. We conduct extensive experiments and achieve new state-of-the-art performances on three relation extraction datasets.
    
    \item We also explore why does \textbf{KLG} work and we find that \textbf{KLG} is particularly good at handling long-tailed classes.
\end{enumerate}

%% file: 3-Method.tex
\section{Our Approach}

\begin{figure*}[t]
	\centering
	\includegraphics[width=0.99\linewidth]{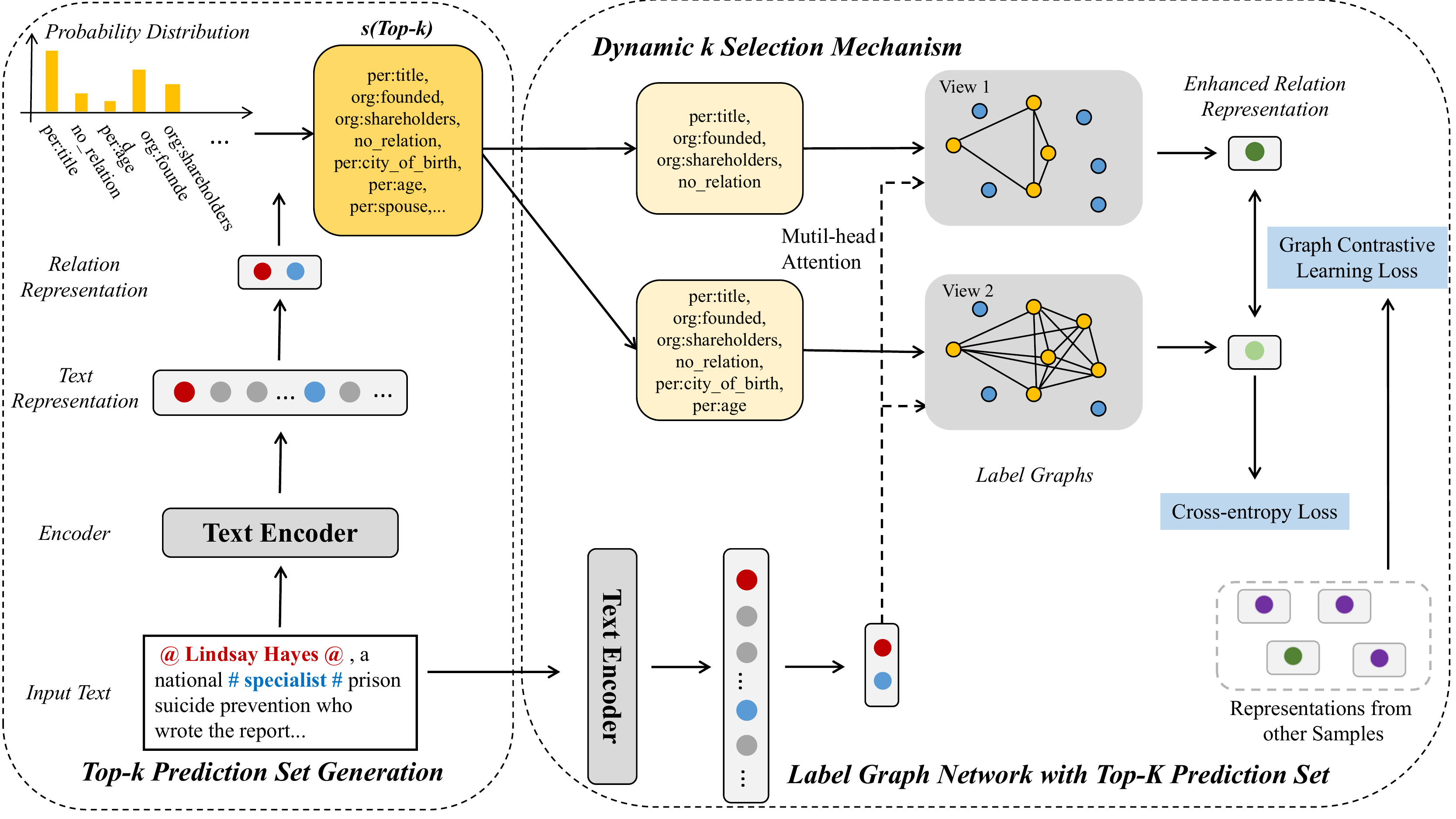}
	\caption{Illustration of our proposed method \textbf{KLG}, the example is taken from TACRED training set. \textbf{Left:} the procedure of \textit{Top-k} prediction set generation. This is a classical supervised learning paradigm. For each sample, instead of only outputting the label with the highest probability, we extract the \textit{Top-k} prediction set depending on the probability distribution. \textbf{Right:} the overall architecture of \textbf{KLG}. We first use pre-defined labels and \textit{s(Top-k)} to build a label graph. Then for each sample, we design a dynamic $k$-selection mechanism to create diverse \textit{Top-k} prediction sets and label graphs, together with graph contrastive learning. \textbf{KLG} could achieve more powerful relation representations. Better viewed in color.}
	\label{model}
\end{figure*}

\subsection{\textit{Top-k} Prediction Set Generation}\label{2.1}
In the first step, we need to generate the \textit{Top-k} prediction set for each sample. In this research, we use a pre-trained PLM to train a baseline model $M$ and obtain the \textit{Top-k} prediction sets. Specifically, for a given input sentence $s$ and two target entities $e_1$ and $e_2$, we first use an entity marker to inclose the entities, and add the entity type information before each target entity. The above two markers highlight the entities and are demonstrated as crucial for relation extraction.\footnote{We only use the entity marker if the entity type information is unavailable, such as the SemEval2010 dataset.} Then we use the max-pooling of the entity spans as the entity representations $\textbf{e}_1$ and $\textbf{e}_2$, where $\textbf{e}_1, \textbf{e}_2 \in \mathbb{R}^{d}$. $d$ is the hidden size of the PLM's output, e.g., 1024 for the RoBERTa-large model. After concatenating the representations of the head entity and the tail entity, we use a single layer to obtain the relation representation $\textbf{r} \in \mathbb{R}^{d}$. 

\begin{equation}
    \textbf{r} = MLP([\textbf{e}_1 || \textbf{e}_2]),
\end{equation}

where $||$ denotes vector concatenation. Finally, a classification layer with a softmax activation function is used to output the probability distribution of all pre-defined labels. 

After we finish the training of baseline model $M$, we use $M$ to predict all samples in the training, dev, and test sets.\footnote{We use the checkpoint that achieves the best results on the dev set as the final baseline model.} For each sample, we generate its \textit{Top-k} prediction set depending on the probability distribution of the classification layer's output, denoted as \textit{s(Top-k)}.

\subsection{Label Graph Fusion}\label{2.2}
In this subsection, we will fine-tune a new PLM by introducing a label graph to leverage the \textit{s(Top-k)} effectively. Typically, a graph $\mathcal{G}$ should have a node set $\mathcal{V}$ and an edge set $\mathcal{E}$, denoted as $\mathcal{G(V, E)}$. The edge links a pair of nodes, denoted as the $(u,v) \in \mathcal{E}$, where $u \in \mathcal{V}$ and $v \in \mathcal{V}$, and all the graphs are undirected, i.e., $(u,v) \in \mathcal{E} \leftrightarrow (v,u) \in \mathcal{E}$. The number of nodes is $N = |V|$. For the node representation, we have a feature matrix $\textbf{X} \in \mathbb{R}^{N*d}$, where $d$ is the feature dimension. For convenience, we set this dimension the same as the hidden size of the PLM's output in \S \ref{2.1}. To represent the edge set $\mathcal{E}$, researchers usually introduce an adjacency matrix $A \in \mathbb{R}^{N*N}$, where $A_{i,j} = 1$ if $(u,v) \in \mathcal{E}$ and $A_{i,j} = 0$ otherwise.

Suppose the current downstream task has $N$ pre-defined labels, denoted as $c_1, c_2, ..., c_N$. We first initialize the label graph with a feature matrix $\textbf{X} \in \mathbb{R}^{N*d}$, and each node representation $\textbf{h}_i$ corresponds to a pre-defined label $c_i$. Where $\textbf{X}$ is random initialized and will be optimized during training. Then for each sample, we construct the edge set $\mathcal{E}$ based on its \textit{s(Top-k)}. We set labels in the \textit{s(Top-k)} are connected, including the self-loop. Specifically, for the node $u$ and node $v$, the edge $(u,v)$ is computed as follows:

\begin{equation}
(u,v) = \begin{cases}
1, c_u, c_v \in \textit{s(Top-k)} \\
0, otherwise, \\
\end{cases}
\end{equation}

For a given \textit{s(Top-k)}, the label graph has $\frac{k(k+1)}{2}$ edges. It is worth noticing that if the ground truth label is not in the \textit{s(Top-k)}, then these edges will not link with the ground truth label in the graph. As we can see in Figure \ref{pilot}, there are still very few samples that are not recalled under \textit{s(Top-10)}.

Then we use graph attention network (GAT) \cite{DBLP:conf/iclr/VelickovicCCRLB18} to process the label graph $\mathcal{G}$. A scoring function $e: \mathbb{R}^{d}*\mathbb{R}^{d} \Rightarrow \mathbb{R}$ computes a score for every edge $(u,v)$, where $A_{i,j} = 1$. $e(\textbf{h}_u, \textbf{h}_v)$ indicates the importance of the features of the neighbor $v$ to the node $u$:

\begin{equation}
    e(\textbf{h}_u, \textbf{h}_v) = LeakyReLU(a^T \cdot [\textit{\textbf{W}}\textbf{h}_u||\textit{\textbf{W}}\textbf{h}_v]),
\end{equation}

where $a \in \mathbb{R}^{2d{'}}$m $\textit{\textbf{W}} \in \mathbb{R}^{d \times d{'}}$ are learned parameters, we set $d^{'} = d$ in this research. After we obtain attention scores across all neighbors of node $u$, a softmax function is used to normalize these scores:

\begin{equation}
    \alpha (u,v) = \frac{exp(e(\textbf{h}_u, \textbf{h}_v))}{\sum_{c_{v^{'}} \in \textit{s(Top-k)}}exp(e(\textbf{h}_u, \textbf{h}_{v^{'}}))}
\end{equation}

Then we compute a weighted average of transformed features of the neighbor node using the normalized attention coefficient as the new node representation:

\begin{equation}
    \textbf{h}^{'}_u = \sigma(\sum_{c_{v} \in \textit{s(Top-k)}} \alpha(u,v) \cdot \textit{\textbf{W}}\textbf{h}_v)
\end{equation}

Then we use a mutil-head attention layer \cite{DBLP:conf/nips/VaswaniSPUJGKP17} to aggregate the relation representation $\textbf{r}$ and those $k$ node representations $\textbf{X}_k$, and $\textbf{r}$ is the query vector:

\begin{equation}
    \hat{\textbf{r}} = MultiHeadAtt(\textbf{r}, \textbf{X}_k, \textbf{X}_k).\label{r}
\end{equation}

Where $\textbf{X}_k$ contains $k$ node representations corresponding to labels in \textit{s(Top-k)}. $\hat{\textbf{r}}$ is the enhanced relation representation that learns label connections from the label graph. Finally, we concatenate the two relation representations and use an MLP layer with a softmax activation function to obtain the final prediction:

\begin{equation}
    p = MLP([\textbf{r} || \hat{\textbf{r}}])
\end{equation}

We use a standard cross-entropy loss to optimize our model, and the loss is denoted as $\mathcal{L}^{CE}$.

\subsection{Dynamic $k$-Selection Mechanism}

While leveraging the label graph could achieve consistent improvements, one crucial problem is determining a proper $k$ in \textit{s(Top-k)}. In \S \ref{2.2}, we select the $k$ where the \textit{Top-k} recall achieves 0.99 on the dev set. However, the performance may sensitive to a fixed $k$ and result in an unstable output. In this subsection, we propose a \textbf{D}ynamic $k$ \textbf{S}election Mechanism (DS), which could create diverse \textit{Top-k} prediction sets for each sample, and further leverages graph contrastive learning for better relation representations. To be specific, we first choose the value of $k$ that the \textit{Top-$k$} recall achieves 0.99 on the dev set. Then we randomly choose another $k^{'}$ from the following uniform distribution:

\begin{equation}
    k^{'} \in [\lceil \frac{1}{2}k \rceil, \lceil \frac{3}{2}k \rceil].\label{ceil}
\end{equation}

Where $\lceil * \rceil$ means round up, and equation \ref{ceil} ensures that we could select diverse $k$ at each training step.\footnote{For example, if k = 5, we will choose k from [3, 8] to build different label graphs}

After applying the dynamic $k$-selection mechanism, we can build multiple label graphs using different \textit{Top-k} prediction sets. In our research, we use \textit{s(Top-k)} and \textit{s(Top-$k^{'}$)} to construct two different label graphs for a given sample, denoted as $\mathcal{G}$ and $\mathcal{G}^{'}$. To fully use these label graphs and achieve more powerful relation representations, we design a graph contrastive learning loss. Specifically, for a given sample $x_{i}$, we first compute two relation representations $\hat{\textbf{r}_i}$ and $\hat{\textbf{r}_i}^{'}$ through Equation \ref{r}. These two relation representations are generated from the specific sample $x_{i}$, thus $\hat{\textbf{r}_i}^{'}$ can be seen as the augmentation version of $\hat{\textbf{r}_i}$. We treat $\hat{\textbf{r}_i}$ and $\hat{\textbf{r}_i}^{'}$ as a positive pair. Additionally, in a mini-batch, samples that hold the same label with $x_{i}$ also can be seen as positives for $\hat{\textbf{r}_i}$, and samples that hold different labels are treated as negative samples. Then for a mini-batch with $N$ training examples, denoted as ${(x_i, y_i)}_{i=1}^N$, we have $2N$ relation representations, where the last half examples of the batch are the augmented views of the first half, and they share the same labels. And $i \in I = [2N]$ is the index of an arbitrary relation representations. Our graph contrastive learning loss is defined as follows:

\begin{equation}
    \mathcal{L}^{GCL} =\sum\limits_{i \in I}\frac{-1}{P(i)}\sum\limits_{j \in P(i)} \mathcal{L}_{i,j}^{GCL}\label{CL2}
\end{equation}

\begin{equation}
    \mathcal{L}_{i,j}^{GCL} =log\frac{exp(\hat{\textbf{r}_i}^T \cdot \hat{\textbf{r}_j}/\tau)}{exp(\hat{\textbf{r}_i}^T \cdot \hat{\textbf{r}_j}/\tau)+\sum\limits_{q \in N(i)}exp(\hat{\textbf{r}_i}^T \cdot \hat{\textbf{r}_{q}}/\tau)}\label{CL1}
\end{equation}

Here $P(i)$ is the positive example set for $\hat{\textbf{r}_i}$, and $N(i)$ is the negative example set for $\hat{\textbf{r}_i}$. $\tau$ is a scalar temperature parameter. Through graph contrastive learning, \textbf{KLG} could effectively utilize diverse label graphs and learn more powerful relation representations.

Finally, the overall training loss for \textbf{KLG} is shown as follows:
\begin{equation}
    \mathcal{L} = \alpha\mathcal{L}^{CE} + (1 - \alpha)\mathcal{L}^{GCL},
\end{equation}

where $\alpha$ is a loss weighting factor. 

%% file: 4-experiments.tex
\section{Experiments}

\subsection{Setting}

\noindent\textbf{Dataset and Metrics.} We evaluate \textbf{KLG} on TACRED \cite{DBLP:conf/emnlp/ZhangZCAM17}, TACRED-Revisit \cite{DBLP:conf/acl/AltGH20a}, and SemEval2010 \cite{DBLP:conf/semeval/HendrickxKKNSPP10}, three commonly used relation extraction datasets. Due to the space limitation, we present detailed information about these datasets in the Appendix \ref{dataset}. Following previous work, we use the Micro-F1 score excluding 'No Relation' as the metric. Besides, we also report the precision, recall, and Macro-F1 score in our detailed analysis.

\noindent\textbf{Model Details.} We use Pytorch \cite{DBLP:conf/nips/PaszkeGMLBCKLGA19} and RoBERTa-large as the text encoder for both parts of \textbf{KLG}. The checkpoint can be downloaded here. \footnote{https://huggingface.co/roberta-large} The batch size is 16, and the optimizer is AdamW \cite{DBLP:conf/iclr/LoshchilovH19} with a 1e-5 learning rate and a warm-up strategy. The maximum training epoch is 10, and the maximum input length is 256. The scalar temperature parameter $\tau$ is 0.05, and the loss weighting factor $\alpha$ is 0.9. We set $k$ as the \textit{Top-$k$} recall achieves 0.99 on the dev set. The checkpoint that achieves the best result on the dev set is used for testing. \footnote{Please refer to Appendix \ref{param} for the detailed information of parameter searching.} We conduct each experiment three times and report the average result to reduce the randomness. Besides, we also report the performances using different backbone networks. Please refer Appendix \ref{othermodel} for more details.

\subsection{Comparison Methods}\label{comparison}

We compare \textbf{KLG} with state-of-the-art RE systems that represent a diverse array of approaches. 

\textbf{Classification-based Methods.} These methods fine-tune PLMs on RE datasets with a standard classification loss. \textbf{LUKE} \cite{DBLP:conf/emnlp/YamadaASTM20} is a popular model for various information extraction tasks, it is pre-trained with large number of external weakly supervised data. \textbf{IRE} \cite{DBLP:journals/corr/abs-2102-01373} is a strong model which uses entity type marker and achieves impressive performances. \textbf{RECENT-SpanBERT} \cite{DBLP:conf/acl/LyuC21} designs multi-task learning for better usage of entity type information.
    
\textbf{Sequence-to-sequence Methods.} These methods use text generation models such as BART \cite{lewis-etal-2020-bart} and T5 \cite{DBLP:journals/jmlr/RaffelSRLNMZLL20} for RE. 
\textbf{REBEL} \cite{DBLP:conf/emnlp/CabotN21} leverages a BART-large model and is pre-trained on a huge external corpus for RE. \textbf{TANL}\cite{DBLP:conf/iclr/PaoliniAKMAASXS21} reformulates RE as a translation task by utilizing the powerful T5 model. Since the above two methods do not leverage entity type information in their original settings. To have a fair comparison, we re-implement these methods with official code and add the entity type information. We also directly treat relation names as generation objectives and fine-tune a \textbf{BART} model on RE datasets.

\textbf{Prompt-based Methods.} This line of research uses prompt and treats RE as a cloze-style task. \textbf{PTR} \cite{DBLP:journals/corr/abs-2105-11259} and \textbf{KnowPrompt} \cite{DBLP:conf/www/ChenZXDYTHSC22} use different prompt and answer word engineering for RE. While \textbf{NLI-DeBERTa} \cite{DBLP:conf/emnlp/SainzLLBA21} reformulates relation extraction as an entailment task.

Besides, we also design several other models for comparison. \textbf{Base Model} is the basic network that used for generating the \textit{Top-k} prediction set in \S \ref{2.1}. Based on the \textbf{Base Model}, we design two improved methods that utilize the \textit{Top-k} prediction set in different ways. \textbf{Base Model(P)} adds a \textbf{P}rompt $P$ after the input text, where $P$ is: \uline{Choose a relation from \textit{s(Top-k)} for $e_1$ and $e_2$}. \textbf{Base Model(LS)} uses \textbf{L}abel \textbf{S}moothing \cite{DBLP:conf/icml/IoffeS15} and modifies the one-hot label as the soft label. 

\subsection{Main Results}\label{main_result}
Table \ref{main} shows the performances on three RE datasets, and we can have the following observations.

\textbf{KLG} outperforms all previous state-of-the-art methods on three RE datasets, including pushing the TACRED F1-score to 75.6\%, TACRED-Revisit F1-score to 84.1\%, and SemEval2010 F1-score to 90.5\%. Compared with the most popular classification-based methods, \textbf{KLG} shows favorable results without any external dataset usage or additional pre-training stages. Although the newly emerging methods offer desirable performances with larger PLMs, such as T5 and DeBERTa, \textbf{KLG} still suppresses them by a large margin. The above results verify the effectiveness of \textbf{KLG}. By utilizing \textit{s(Top-k)} via a label graph and the dynamic $k$-selection mechanism, \textbf{KLG} could fully use the available information existing in the \textit{Top-k} prediction set and achieve better performance.
    
Although we apply different technologies to utilize the \textit{Top-k} prediction set, the performances of Base Model(P) and Base Model(LS) are undesirable. Base Model(P) shows that directly injecting the \textit{Top-k} prediction set into the input text can not achieve better performance as we expected, even with the help of a prompt. We think there are two possible reasons for the above phenomenon: 1) the backbone network can not understand the semantic meaning of relation names, and appending the \textit{Top-k} prediction set after the input text may confuse the model's encoder, making the model pays less attention to the original input text. 2) With sufficient training samples, models may learn to select a relation name from the \textit{Top-k} prediction set directly without considering the original input text. As for Base Model(LS), it obtains slight improvements compared with its corresponding baseline model. Meanwhile, \textbf{KLG} achieves much better performances than the above methods. 
    
We also report the ablation study on two crucial components of \textbf{KLG}: 1) the usage of label graph with \textit{Top-k} prediction set, and 2) the dynamic $k$-selection mechanism. The results show that both parts bring notable improvements over the Base Model. With the help of the label graph, our model could review candidate labels existing in the \textit{top-k} prediction set and learn useful information from them. In addition, after being equipped with the dynamic $k$-selection mechanism, \textbf{KLG} learns more powerful relation representations and further improves its performance.

\begin{table}[]
\centering
\setlength{\tabcolsep}{1.3mm}
\begin{tabular}{cccc}
\toprule[1.5pt]
\multicolumn{1}{l|}{}                         & \multicolumn{1}{l|}{\textbf{TACRED}}     & \multicolumn{1}{l|}{\textbf{TACRED-Rev}} & \multicolumn{1}{l}{\textbf{SemEval}} \\ \hline
\multicolumn{4}{l}{\textit{\textbf{Classification-based Methods}}}                                                                                                                 \\ \hline
\multicolumn{1}{c|}{\textbf{LUKE}}            & \multicolumn{1}{c|}{72.7}                & \multicolumn{1}{c|}{80.8}                    & 90.1                                    \\
\multicolumn{1}{c|}{\textbf{IRE}}             & \multicolumn{1}{c|}{74.6}                & \multicolumn{1}{c|}{83.2}                    & 89.8                                     \\
\multicolumn{1}{c|}{\textbf{RECENT}} & \multicolumn{1}{c|}{\underline{75.2}}          & \multicolumn{1}{c|}{\underline{83.4}}              & -                                        \\ \hline
\multicolumn{4}{l}{\textit{\textbf{Sequence-to-Sequence Methods}}}                                                                                                                 \\ \hline
\multicolumn{1}{c|}{\textbf{BART}}           & \multicolumn{1}{c|}{72.7}                & \multicolumn{1}{c|}{81.5}                       & \multicolumn{1}{c}{89.5}                                        \\
\multicolumn{1}{c|}{\textbf{REBEL}}           & \multicolumn{1}{c|}{73.7}                & \multicolumn{1}{c|}{-}                       & -                                        \\
\multicolumn{1}{c|}{\textbf{TANL}}            & \multicolumn{1}{c|}{74.8}                & \multicolumn{1}{c|}{-}                       & -                                        \\ \hline
\multicolumn{4}{l}{\textit{\textbf{Prompt-based Methods}}}                                                                                                                         \\ \hline
\multicolumn{1}{c|}{\textbf{PTR}}             & \multicolumn{1}{c|}{72.4}                & \multicolumn{1}{c|}{81.4}                    & \multicolumn{1}{c}{89.9}                                       \\
\multicolumn{1}{c|}{\textbf{KnowPrompt}}      & \multicolumn{1}{c|}{72.4}                & \multicolumn{1}{c|}{82.4}                    & \multicolumn{1}{c}{\underline{90.2}}                                        \\
\multicolumn{1}{c|}{\textbf{NLI-DeBERTa}}     & \multicolumn{1}{c|}{73.9}                & \multicolumn{1}{c|}{-}                       & -                                        \\ \hline
\multicolumn{4}{l}{\textit{\textbf{Ours}}}                                                                                                                                         \\ \hline
\multicolumn{1}{c|}{\textbf{Base Model}}             & \multicolumn{1}{c|}{74.3} & \multicolumn{1}{c|}{83.1}     & 89.6                      \\
\multicolumn{1}{c|}{\textbf{Base Model(P)}}             & \multicolumn{1}{c|}{74.1} & \multicolumn{1}{c|}{82.7}     & 89.1                      \\
\multicolumn{1}{c|}{\textbf{Base Model(LS)}}             & \multicolumn{1}{c|}{74.6} & \multicolumn{1}{c|}{83.5}     & 89.9                      \\
\multicolumn{1}{c|}{\textbf{KLG}}             & \multicolumn{1}{c|}{\textbf{75.6(+0.4)}} & \multicolumn{1}{c|}{\textbf{84.1(+0.7)}}     & \textbf{90.5(+0.3)}                      \\ 
\multicolumn{1}{c|}{\textbf{KLG w/o DS}}             & \multicolumn{1}{c|}{75.0} & \multicolumn{1}{c|}{83.6}     & 90.1                      \\
\bottomrule[1.5pt]
\end{tabular}
\caption{Micro-F1 score of test sets on three relation extraction datasets. Results are all cited from published papers or re-implemented using official code. For each dataset, \underline{underlined} indicates previous state-of-the-art, \textbf{bold} indicates the best model performance. Results of our methods are averaged over three random seeds, and the results are statistically significant with $p < 0.05$.}
\label{main}
\end{table}


\begin{figure}[h]
	\centering
	\includegraphics[width=0.98\linewidth]{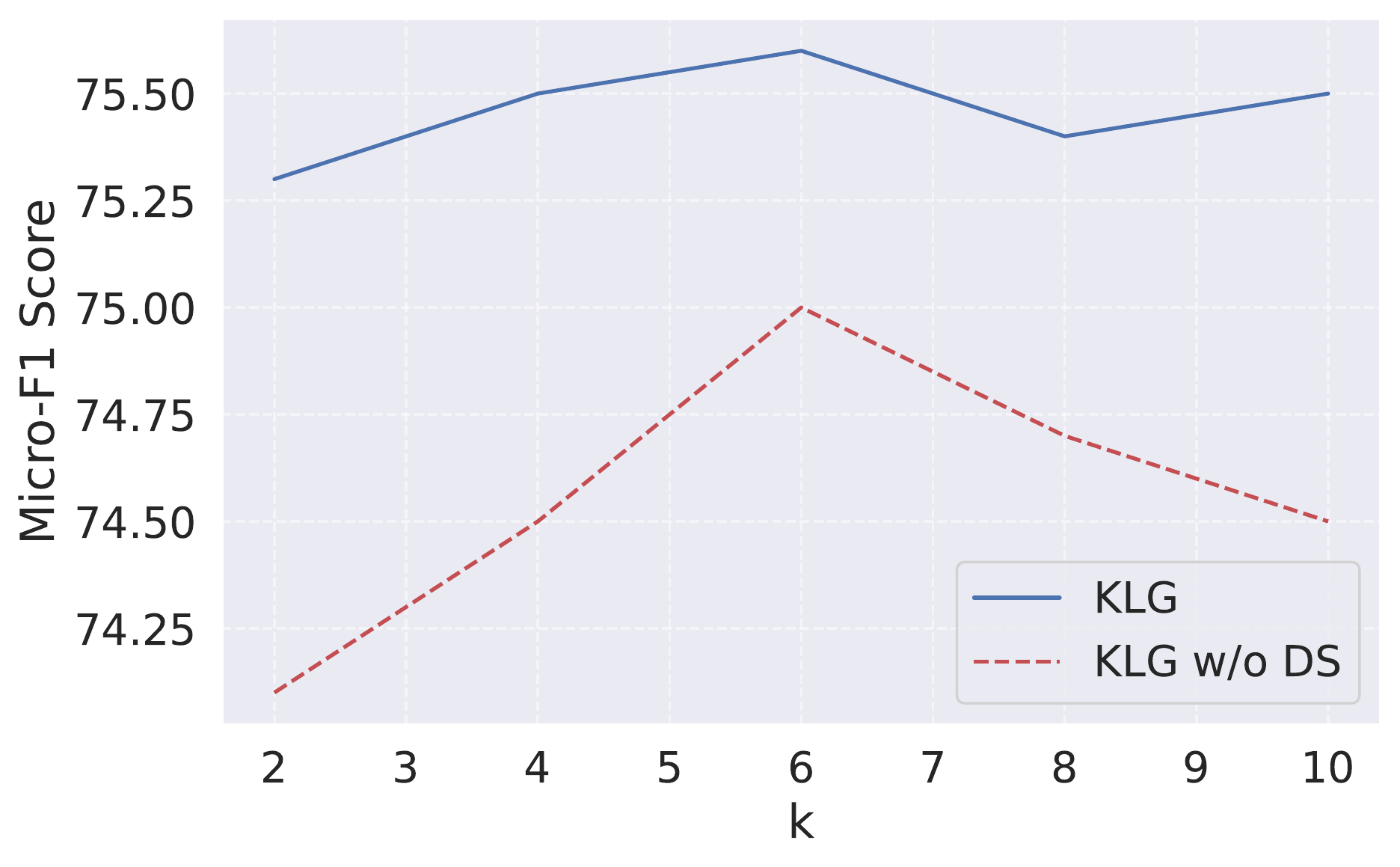}
	\caption{The effectiveness of $k$. We find that the fixed $k$ results in unstable performances. With the help of the dynamic $k$-selection mechanism, \textbf{KLG} could achieve desirable results across a wide range of $k$.}
	\label{k_number}
\end{figure}

\subsection{The Sensitivity of $k$}
In this section, we explore the sensitivity of $k$. These results also verify the robustness of our dynamic $k$-selection mechanism (DS). We present the performance on TACRED test set with DS and without DS in Figure \ref{k_number}. We find that without DS, the performances are very sensitive to the $k$-selection, especially when $k$ is too small or too large. With the help of DS, \textbf{KLG} could achieve more stable and much better performances across a wide range of $k$.


%% file: 5-analysis.tex
\section{Analysis}
\begin{table*}[h]
\begin{tabular}{c|c|c|c|c}
\hline
\textbf{Input Sentence}                                                                                                                                   & \textbf{Ground Truth}                                                 & \textbf{Base Model} & \textbf{Top-6 Prediction Set}                                                                                                                                                      & \textbf{KLG}                                                          \\ \hline
\begin{tabular}[c]{@{}c@{}}\textcolor{blue}{PDA}'s advisory board includes \\ seven members of congress and\\  ..., \textcolor{blue}{Thom Hartmann} rev. \\ lennox yearwood ....\end{tabular} & \begin{tabular}[c]{@{}c@{}}org:top\_members/\\ employees\end{tabular} & org:shareholders    & \begin{tabular}[c]{@{}c@{}}org:shareholders, no\_relation, \\ \textcolor{red}{org:top\_members/employees}, \\ org:founded\_by, org:parents, \\ org:subsidiaries\end{tabular}                         & \begin{tabular}[c]{@{}c@{}}org:top\_members/\\ employees\end{tabular} \\ \hline
\begin{tabular}[c]{@{}c@{}}\textcolor{blue}{She} testified that she had\\  been forced to sign ..., \\ \textcolor{blue}{FBI} price and wood, under \\ threats of ...\end{tabular}             & per:employee\_of                                                      & no\_relation        & \begin{tabular}[c]{@{}c@{}}no\_relation, \textcolor{red}{per:employee\_of}, \\ per:schools\_attended,\\ per:cities\_of\_residence,\\ per:other\_family, \\ per:countries\_of\_residence\end{tabular} & per:employee\_of                                                      \\ \hline
\begin{tabular}[c]{@{}c@{}}He was later flown back to \\ \textcolor{blue}{Manila}, where \textcolor{blue}{he} reunited \\ with family and friends .\end{tabular}                              & \begin{tabular}[c]{@{}c@{}}per:cities\_of\_\\ residence\end{tabular}  & per:city\_of\_birth & \begin{tabular}[c]{@{}c@{}}per:city\_of\_birth, no\_relation, \\ per:countries\_of\_residence, \\ \textcolor{red}{per:cities\_of\_residence}, \\ per:city\_of\_death, per:religion\end{tabular}     & \begin{tabular}[c]{@{}c@{}}per:cities\_of\_\\ residence\end{tabular}  \\ \hline
\end{tabular}
\caption{Case Study. All examples are extracted from the test set of TACRED. \textcolor{blue}{Blue} denotes the target entity, and \textcolor{red}{Red} denotes the ground truth label existing in the \textit{Top-6} prediction set.}
\label{case}
\end{table*}

\subsection{Why Does \textbf{KLG} Work?}\label{long_tail}

\begin{table}[]
\centering
\setlength{\tabcolsep}{2mm}
\begin{tabular}{c|c|c|c}
\toprule[1.5pt]
                                                                               &                    & \textbf{Base Model} & \textbf{KLG} \\ \hline
\multirow{3}{*}{\textbf{\begin{tabular}[c]{@{}c@{}}Head Classes\\ (69.6\%)\end{tabular}}} & \textbf{Precision} & 76.0                & 78.8(+2.8)   \\
                                                                               & \textbf{Recall}    & 79.7                & 77.3(-2.4)   \\
                                                                               & \textbf{Micro-F1}  & 77.8                & 78.1(+0.3)   \\ \hline
\multirow{3}{*}{\textbf{\begin{tabular}[c]{@{}c@{}}Tail Classes\\ (3.6\%)\end{tabular}}} & \textbf{Precision} & 74.4                & 75.3(+0.9)   \\
                                                                               & \textbf{Recall}    & 69.6                & 74.7(+5.1)   \\
                                                                               & \textbf{Micro-F1}  & 71.9                & 75.0(+3.1)   \\ \hline
\multirow{2}{*}{\textbf{\begin{tabular}[c]{@{}c@{}}All Classes\end{tabular}}}  & \textbf{Micro-F1}  & 74.3                & 75.6(+1.3)   \\
                                                                               & \textbf{Macro-F1}  & 61.2                & 65.2(+4.0)   \\ \bottomrule[1.5pt]
\end{tabular}
\caption{Performances of TACRED test set on head classes, tail classes, and all classes. To have a clear picture, we report the precision, recall, Micro-F1 score, and Macro-F1 score. After filtering out the 'No Relation' class, we choose 10 classes with the highest frequency as head classes, and 10 other classes with the lowest frequency are long-tailed classes.}
\label{classes}
\end{table}

As \textbf{KLG} could learn label connections from the \textit{Top-k} prediction set and reconsider these labels, which may particularly benefit long-tailed classes. We want to explore why does \textbf{KLG} work from the perspective of long-tailed classes. We report fine-grained experimental results on the test set of TACRED. After filtering out the 'No Relation' class, we choose 10 classes with the highest frequency as head classes, and 10 other classes with the lowest frequency are long-tailed classes. The proportion of head classes samples is 69.6\%, and only 3.6\% for long-tailed classes. We report the precision, recall, and Micro-F1 score on both head and long-tailed classes. As the Macro-F1 score is more suitable for the imbalanced scenario, we also report the Macro-F1 score on the whole dataset. We can draw following conclusions from Table \ref{classes}.


Both models have undesirable performances on the long-tailed classes than on the head classes. This indicates that dealing with long-tailed classes is more challenging. As for the performance gap between these two types of classes, \textbf{KLG} is much smaller than Base Model, i.e., -3.1\% v.s. -5.9\%. Compared with the Base Model, \textbf{KLG} achieves a remarkable improvement in terms of Macro-F1 score, i.e., 4.0 point absolute improvement. The above results show that \textbf{KLG} is particularly good at handling imbalanced scenario.
    
Considering the performances of long-tailed classes, the improvement of \textbf{KLG} over the Base Model is more higher on long-tailed classes than on head classes (+3.1\% v.s. +0.3\%). Furthermore, \textbf{KLG} dramatically improves the recall from 69.6\% to 74.7\%, with around 5.1 points absolute improvement. We attribute this to the usage of the \textit{Top-k} prediction set and the dynamic $k$-selection mechanism. As the \textit{Top-k} prediction set may contain candidate long-tailed classes which are ignored by previous methods, \textbf{KLG} could review these labels and finally benefit the performance of long-tailed classes. These results show that the performance improvements are mainly from the long-tailed classes. 
    
As for the head classes, although the Micro-F1 score has slight improvement, we find that the precision improves over 2.8\%, since \textbf{KLG} mitigates the trend of false positive on the head classes by paying more attention to the long-tailed classes. Besides, we also observe the recall of head classes has a significant degradation. One possible reason is that \textbf{KLG} tends to classify head classes as long-tailed classes, resulting in a relatively low recall of head classes.

\begin{figure}[h]
	\centering
	\includegraphics[width=0.98\linewidth]{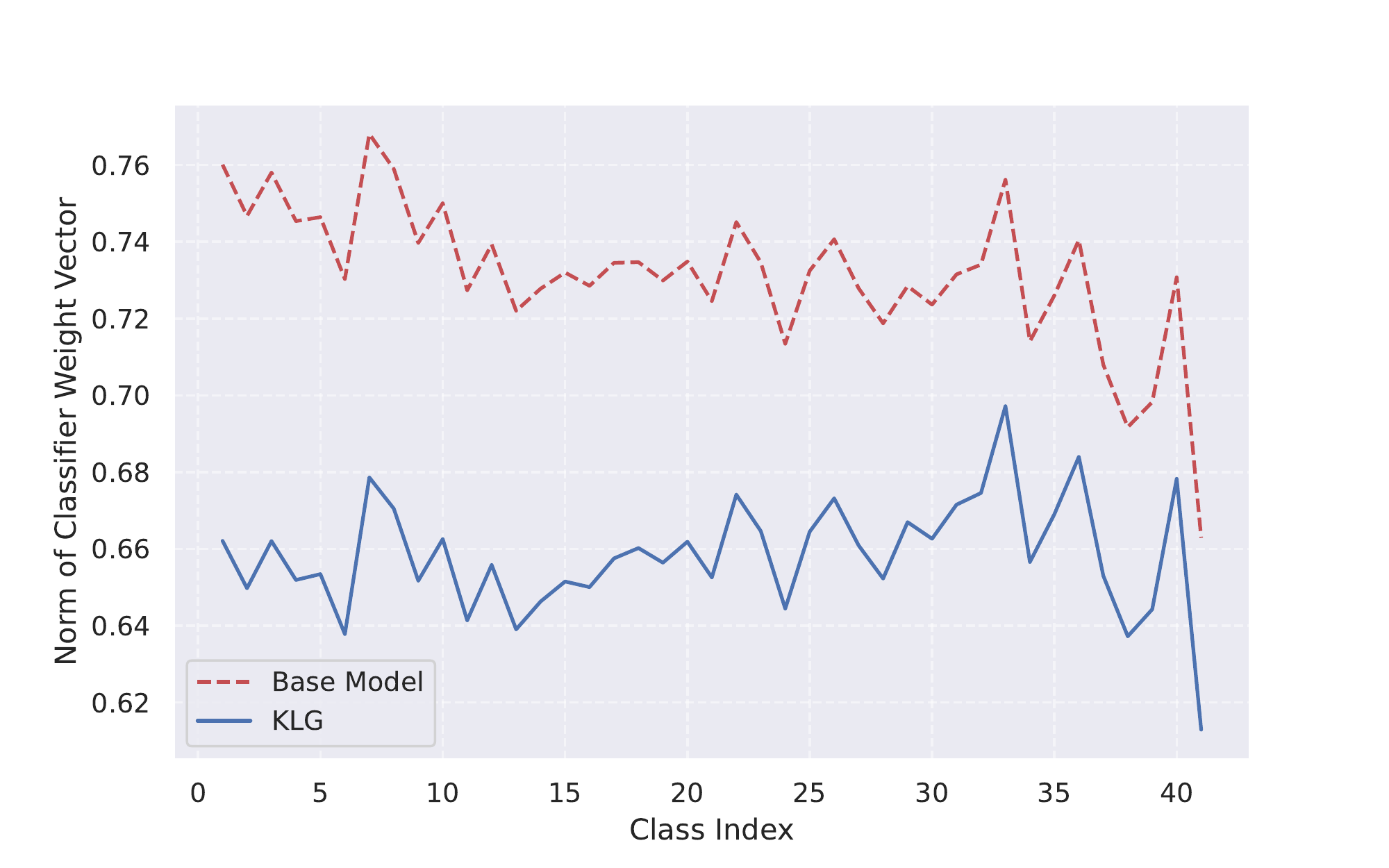}
	\caption{The norm of the classifier weight vector for each class. The class indexes are sorted by the number of samples (head to tail). We exclude the 'No Relation' to better view the trend. We can observe that \textbf{KLG} has similar vector norms across all classes.}
	\label{norm}
\end{figure}

We also conduct visualization to verify the above conclusion intuitively. Previous works \cite{DBLP:conf/iclr/KangXRYGFK20,DBLP:journals/corr/abs-2110-10048,DBLP:conf/mm/PengHGZJY21} show that the $l2$ norm of the classifier weight vector is positively correlated with the number of training samples of the corresponding label. Having a high weight norm for a class means that the classifier tends to output a high logit score for this class, so a good classifier should have essentially the same vector weights. Figure \ref{norm} shows the weight norm of the Base Model and \textbf{KLG}. The number of samples is reduced from the left (head classes) to the right (long-tailed classes). We can observe that the weight norms of the Base Model are significantly different for different classes. On the other hand, \textbf{KLG} has a more balanced weight norm compared with the Base Model, thus is more friendly for long-tailed classes and finally achieves remarkable improvements.

\subsection{Case Study}
To have a clear picture of the effectiveness of \textbf{KLG}, we present several cases in Table \ref{case}. From the first and third cases, we can see that although the Base Model outputs the wrong prediction, the prediction is related to the ground truth label. For example, in the first case, the ground truth is \textit{org:top\_members\/employees}, while the label with the highest probability is \textit{org:shareholders}, which has a similar semantic meaning with \textit{org:top\_members\/employees}. We find the \textit{Top-6} prediction set contains the ground truth label, but this information was ignored by previous works. As for the second case, although Base Model gives \textit{per:employee\_of} a high probability, it finally outputs \textit{'no\_relation'} as the final prediction. Since the majority of examples in the training set are labeled as \textit{'no\_relation'}, Base Model tends to output a high logit score for \textit{'no\_relation'} class. \textbf{KLG} successfully outputs correct predictions in all examples by reviewing candidate labels and learning useful information from the \textit{Top-k} prediction set. 

%% file: 2-related_work.tex
\section{Related Work}

Relation extraction is a well-studied and popular task in natural language processing. Early works leveraged machine learning to tackle relation extraction, and these methods can be divided into two categories: feature-based methods \cite{kambhatla2004combining,boschee2005automatic,chan2010background,sun2011semi,Nguyen2014Regularization} and kernel-based methods \cite{Culotta2004Dependency,zhou2007tree,giuliano2007fbk,qian2008exploiting,Nguyen2009Structures,Sun2014tree}. Feature-based methods heavily rely on handcraft features. These features are domain-specific and data-specific. Kernel-based methods need existing NLP tools to transform an input sentence into a parse tree. However, the NLP tools may make mistakes during processing, which may negatively affect the model's performance.

Then deep learning methods have been the mainstream solutions for RE. \citet{liu2013convolution} first adopt deep learning for RE via convolutional neural network. \citet{zeng2015distant} proposed a model named Piecewise Convolutional Neural Networks(PCNN). Also, there are lots of works modified convolutional neural network and recurrent neural network for RE \cite{santos2015classifying,miwa2016end,zhou2016attention,she2018distant,he2018see,su2017global}. 

Recently, transformer-based methods which leverage PLMs have shown impressive performances on RE \cite{DBLP:conf/emnlp/YamadaASTM20,DBLP:conf/aaai/XueSZC21,DBLP:conf/coling/LiYSXXZ20,DBLP:conf/aaai/LiYHZ21,DBLP:conf/emnlp/RoyP21,DBLP:journals/corr/abs-2102-01373,DBLP:conf/acl/LyuC21}. These methods used the PLMs such as BERT and Roberta as backbone network, and designed novel components or multi-task learning framework for better performance. 

Prompt-based methods \cite{radford2019language} were drawn some attention in recent research as well. PTR \cite{DBLP:journals/corr/abs-2105-11259} and KnowPrompt \cite{DBLP:conf/www/ChenZXDYTHSC22} used a human-designed prompt with different label verbalizers for each relation. While NLI-DeBERTa \cite{DBLP:conf/emnlp/SainzLLBA21} reformulated relation extraction as an entailment task with relation descriptions. Sequence-to-sequence methods are another interesting research direction for RE. REBEL \cite{DBLP:conf/emnlp/CabotN21} focused on joint entity and relation extraction and pre-trained a BART-large model with external datasets. TANL \cite{DBLP:conf/iclr/PaoliniAKMAASXS21} used T5 and backbone network and achieved impressive performance on various information extraction tasks. 

In this research, we focus on leveraging the \textit{top-k} prediction set for RE, which is not explored in past work. 